\documentclass[10pt,a4paper]{article}

\usepackage[T1]{fontenc}
\usepackage[utf8]{inputenc}
\IfFileExists{XCharter.sty}{\usepackage{XCharter}}{\usepackage{charter}}
\usepackage{microtype}
\usepackage[
  a4paper,
  left=22mm,right=22mm,
  top=29mm,bottom=32mm,
  headheight=12pt,headsep=16pt,
  footskip=30pt
]{geometry}
\usepackage{setspace}
\usepackage{amsmath,amssymb}
\usepackage{graphicx}
\usepackage{booktabs,tabularx}
\usepackage{enumitem}
\usepackage{xcolor}
\usepackage{amsmath}
\DeclareMathOperator*{\argmin}{arg\,min}
\usepackage{mathtools}
\usepackage{makecell}
\usepackage{wrapfig}
\usepackage{adjustbox}
\usepackage{siunitx}
\usepackage{algorithm}
\usepackage{algpseudocode}
\usepackage{amssymb}
\usepackage{array}
\usepackage{wrapstuff}

\newcommand{\val}[2]{#1\,{\scriptsize\textcolor{gray!95}{$\pm$\,#2}}}

\usepackage{pifont}
\let\oldding\ding
\renewcommand{\ding}[2][1]{\scalebox{#1}{\oldding{#2}}}

\definecolor{accent}{HTML}{76B900}
\definecolor{urlblue}{HTML}{000080}
\definecolor{rulegray}{gray}{0.50}
\definecolor{placeholderbg}{HTML}{F5F6F3}
\definecolor{placeholdertext}{HTML}{687260}
\usepackage{caption}
\usepackage{subcaption}
\setlist{leftmargin=*,topsep=4pt,itemsep=2pt,parsep=0pt}
\usepackage[numbers,sort&compress,round]{natbib}
\setcitestyle{numbers,round,semicolon}
\usepackage[colorlinks=true,linkcolor=accent,citecolor=accent,
  urlcolor=urlblue,bookmarksnumbered=true]{hyperref}
\newcommand{\papertitle}{Looking Back to Move Forward: \\ Temporal Verification for Generative Robot Policies}
\newcommand{\shorttitle}{Temporal Verification for Generative Robot Policies}
\newcommand{\paperauthors}{%
  Haoxuan Wang\textsuperscript{1},
  Wayne Wu\textsuperscript{2},
  Yan Yan\textsuperscript{1,\(\dagger\)},
  Bolei Zhou\textsuperscript{2,\(\dagger\)}%
}
\newcommand{\paperaffiliations}{%
  \textsuperscript{1}University of Illinois Chicago
  \quad
  \textsuperscript{2}University of California, Los Angeles \quad 
  \textsuperscript{\(\dagger\)}Equal advising
}
\newcommand{\paperurl}{https://hatchetproject.github.io/tev/}
\newcommand{\paperurltext}{Project Page}
\newcommand{\covernote}{}
\hypersetup{pdftitle={Research Paper Template},pdfauthor={}}

\usepackage{fancyhdr}
\renewcommand{\headrule}{\hbox to\headwidth{\color{rulegray}\leaders\hrule height 1pt\hfill}}
\renewcommand{\footrule}{\hbox to\headwidth{\color{rulegray}\leaders\hrule height 1pt\hfill}}

\fancypagestyle{cover}{%
  \fancyhf{}
  \renewcommand{\headrule}{}%
  \fancyfoot[L]{\fontsize{8}{10}\selectfont\covernote}
}
\usepackage{titlesec}
\titleformat{\section}{\fontsize{13}{16}\selectfont\bfseries}{\thesection.}{0.65em}{}
\titleformat{\subsection}{\fontsize{12}{15}\selectfont\bfseries}{\thesubsection.}{0.65em}{}
\titleformat{\subsubsection}{\normalsize\bfseries}{\thesubsubsection.}{0.65em}{}
\titleformat{\paragraph}[runin]{\normalsize\bfseries}{}{0pt}{}
\titlespacing*{\section}{0pt}{18pt plus 2pt minus 2pt}{8pt}
\titlespacing*{\subsection}{0pt}{12pt plus 2pt minus 2pt}{6pt}
\titlespacing*{\subsubsection}{0pt}{10pt}{4pt}
\titlespacing*{\paragraph}{0pt}{6pt}{0.9em}
\renewenvironment{abstract}{%
  \begingroup
  \setlength{\parskip}{0pt}%
  \titlespacing*{\section}{0pt}{14pt}{4pt}%
  \section*{Abstract}\setstretch{1.2}\bfseries\noindent
}{\par\endgroup}
\newcommand{\keywords}[1]{%
  \par\noindent{\fontsize{11}{13}\selectfont\itshape Keywords: #1}\par
}

\newcommand{\figureplaceholder}[2]{%
  \begingroup
  \setlength{\fboxsep}{0pt}%
  \colorbox{placeholderbg}{%
    \parbox[c][#1][c]{\linewidth}{%
      \centering\color{placeholdertext}\normalfont\small #2\par
    }%
  }%
  \endgroup
}

\newcommand{\makepaperheading}{%
  \begingroup
  \setstretch{1}%
  \setlength{\parindent}{0pt}%
  \setlength{\parskip}{0pt}%
  \raggedright
  {\color{black}\fontsize{17}{21}\selectfont\bfseries\papertitle\par}%
  \vspace{12pt}%
  {\fontsize{9}{12}\selectfont\bfseries\paperauthors\par}%
  \vspace{3pt}%
  {\fontsize{8}{10}\selectfont\paperaffiliations\par}%
  \vspace{3pt}%
  {\fontsize{8}{10}\selectfont
    \href{\paperurl}{\texttt{\paperurltext}}\par}%
  \endgroup
}

\newif\ifcoverpage
\coverpagefalse

\begin{document}

\ifcoverpage
\else
  \thispagestyle{cover}
  \makepaperheading
\fi

\begin{abstract}
Generative policies have emerged as a promising paradigm for robot learning, combining expressive generative action modeling with scalable imitation learning from large demonstration corpora. However, heterogeneous demonstrations can induce suboptimal action chunks whose errors compound over time, eventually driving the robot into out-of-distribution states from which recovery is difficult. Action verification offers a test-time scaling strategy for mitigating this failure mode by sampling multiple candidate actions and using a verifier to select one for execution. Existing approaches, however, remain temporally myopic and costly to train, evaluating candidates from the current observation alone without accounting for trajectory continuity and often relying on large verifiers and additional expert demonstrations. In this paper, we introduce \textbf{Te}mporal \textbf{V}erification (\textbf{TeV}), an efficient temporally aware action verification framework for flow-matching VLAs. TeV first learns a \emph{temporal token} that summarizes recent observation--action history, enabling candidate chunks to be evaluated as continuations of the execution trajectory rather than as isolated predictions. Conditioned on this token, TeV constructs positive--negative pairs without additional expert demonstrations or preference annotations and trains an energy-based verifier contrastively to assign lower energy to higher-quality, trajectory-consistent action chunks. Beyond post-hoc ranking, TeV further uses the learned energy landscape to guide intermediate flow samples toward lower-energy regions, improving candidates before final selection. Extensive experiments in simulation and real-world settings demonstrate that TeV provides reliably ranks action candidates, improves task success rates, and produces smoother execution trajectories.
\end{abstract}

\section{Introduction}
Generative policies based on diffusion and flow models have rapidly become a prominent paradigm in robot learning~\citep{diffusion_policy,flow_matching,pi0}. By scaling imitation learning to large corpora of human demonstrations, these models have enabled increasingly complex robotic skills, including household cleaning~\citep{pi05}, cloth folding~\citep{kai0}, and contact-rich assembly such as screw driving~\citep{rlt}. Among these approaches, Vision-Language-Action (VLA) policies trained with flow-matching objectives~\citep{flow_matching} have emerged as a promising framework for generalist robot control~\citep{pi05,smolvla,gr00t,lingbot_vla,intern_vla}. These models combine semantic representations from vision-language backbones with continuous control, mapping language instructions, visual observations, and robot states to temporally extended action chunks. Flow matching further provides a tractable training objective with efficient sampling, making it well suited to scalable robot learning.

Despite this promise, the flexibility of flow-matching policies also introduces a critical vulnerability. When trained on heterogeneous demonstrations, these policies can generate suboptimal action chunks. Small errors can then compound over time, driving the robot away from the demonstrated state--action manifold and into out-of-distribution states from which recovery becomes increasingly difficult. Existing approaches mitigate this issue by retraining or fine-tuning the base policy through reinforcement learning~\citep{rl_flow,residual_rl,robustvla} or representation regularization~\citep{dig_flow,rs_cl}. While effective, these approaches incur substantial training cost and risk degrading the generalist capabilities encoded by the base policy. Verification-based test-time scaling~\citep{eve,taco} therefore offers an alternative, improving action selection at deployment time without modifying the base policy. Rather than committing to a single action chunk, these methods use additional inference-time compute to sample multiple candidates from the base policy. A separate verifier then scores the candidates according to learned expert or preference signals and selects the highest-ranked action for execution.

Yet existing verification-based methods face two key limitations. First, they suffer from temporal myopia, generating and evaluating candidate actions based solely on the current observation. 
For generative policies, successful execution requires not only selecting high-quality candidates for the current observation, but also ensuring that consecutive action chunks form a coherent trajectory. Without cross-chunk consistency, the policy may oscillate between individually plausible modes, resulting in jerky motion, unstable execution, and task failure~\citep{bid,remac}. 
Effective verification must therefore account for both \textit{local action plausibility} and \textit{temporal consistency} with recent execution history. 
Second, existing verifier designs impose substantial training and deployment overhead. 
Existing approaches often rely on large VLM-based verifiers~\citep{robomonkey,eve}, requiring careful task adaptation and substantial additional supervision. Obtaining such supervision can be expensive, as reward labels, trajectory preferences, or failure traces require additional annotation, rollout collection, or real-world interaction~\citep{dagger,have}. Large verifiers further complicate real-world deployment by adding inference overhead to latency-sensitive control loops, particularly under limited onboard compute~\citep{vla_perf}.

To address these limitations, we propose \textbf{Te}mporal \textbf{V}erification (\textbf{TeV}), an efficient energy-based verification framework for flow-matching VLAs. TeV improves both action quality and temporal consistency by combining post-hoc candidate ranking with energy-guided generation. Specifically, we first construct a \emph{temporal token} that summarizes the robot's recent observation--action history. 
Encoded efficiently and regularized with a dynamics-aware objective, the token provides the verifier with execution context beyond the current observation. Conditioned on this token, we instantiate the verifier as a lightweight energy model that scores sampled action chunks and ranks them for execution. Because candidate selection requires only relative comparison rather than calibrated reward prediction, we train the verifier contrastively using expert demonstration chunks as positives and temporally mismatched or policy-generated chunks as informative negatives. Finally, TeV extends verification beyond post-hoc ranking by using gradients of the learned energy function to guide flow integration, guiding intermediate samples toward lower-energy regions before final selection.

TeV is a lightweight add-on for flow-matching VLAs that evaluates sampled action chunks against recent execution history rather than the current observation alone.
It increases the parameter count by less than 0.15\%, requires no additional expert demonstrations, and is the first action-verification framework to integrate candidate ranking with generation-time guidance.
Extensive simulation and real-world experiments show that TeV reliably ranks action candidates, improves task success by 6--18 percentage points, and produces smoother execution trajectories, demonstrating the benefits of temporally aware verification for generative robot control.

\section{Related Work}
\textbf{Test-time Scaling via Action Verification.}
Test-time verification improves model performance by allocating additional inference-time computation to evaluate and rank candidate outputs~\citep{llm_verify}. In robotics, this paradigm has been explored through different verifier designs. External-verifier methods use vision-language models to evaluate candidate actions or plans~\citep{robomonkey,eve}. RL-based approaches support verification through online policy adaptation~\citep{have} or learned value functions for action ranking~\citep{steer_rl}. Other methods use verification to improve instruction following~\citep{cover}. The most relevant verification approach to ours is TACO~\citep{taco}, which uses a pseudo-count estimator to favor candidate actions near high-density successful modes of the training distribution. However, existing methods rely on static timestep observations, whereas our method conditions verification on recent execution history to encourage temporally coherent and smooth behavior.

\noindent\textbf{Temporal Consistency for Chunked Policies.}
Flow-matching policies can generate locally plausible yet temporally inconsistent action chunks, causing mode oscillation and jerky behaviors. Existing methods address this by modifying the chunk execution process. ACT~\citep{aloha} smooths execution by averaging overlapping action chunks, while BID~\citep{bid} samples multiple candidates and selects the chunk most consistent with the previous prediction over overlapping timesteps. Real-time control methods~\citep{rtc,vlash,remac} study temporal consistency under asynchronous execution, where stale observations further complicate the problem. In contrast, our method learns a history-conditioned energy verifier that evaluates whether each candidate is an expert-compatible continuation of the recent observation–action trajectory.

\begin{figure}[t]
    \centering
    \includegraphics[width=1.0\linewidth, trim=160 175 210 145, clip]{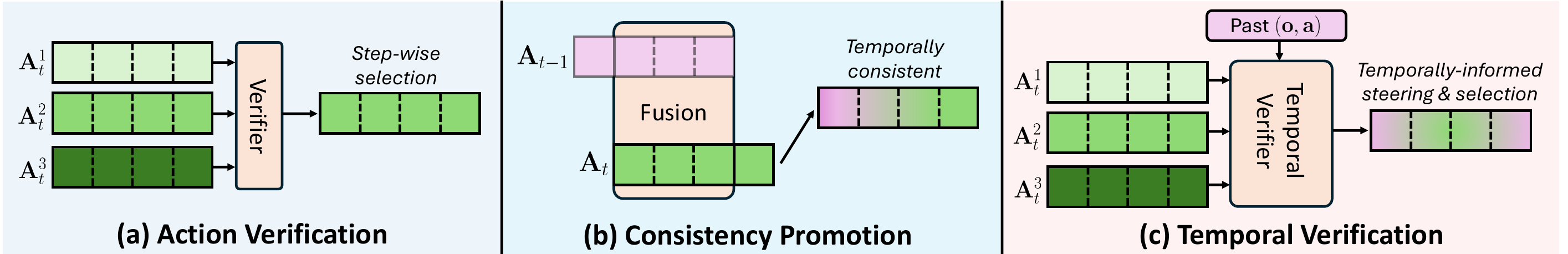}
    \vspace{-12pt}
    \captionsetup{
        justification=raggedright,
        singlelinecheck=false
    }
    \caption{\textbf{Paradigm comparison.}
    \textbf{(a)} Conventional action verification selects among multiple sampled candidates at each decision step, relying only on the current observation.
    \textbf{(b)} Temporal-consistency methods refine the current prediction by using previous action chunks to promote cross-chunk coherence.
    \textbf{(c)} Temporal Verification integrates action verification with temporal consistency modeling. Conditioned on past observation-action context, the learned temporal verifier guides and selects action chunk candidates that are more consistent with recent execution.
    }
    \label{fig:compare}
\end{figure}

\noindent\textbf{Policy Steering.}
Inference-time steering improves pretrained generative policies by modifying the sampling or refinement process while keeping the base policy fixed~\citep{steer_rl}. Existing methods typically use external objectives or auxiliary models to guide generated actions toward desired task, safety, or user-specified constraints. Model-predictive refinement methods~\citep{mpc_safety,omniguide} guide policy outputs using learned dynamics models to satisfy safety constraints or improve task performance, while human-in-the-loop approaches~\citep{human_loop_1,human_loop_3} refine actions based on user-provided subgoals, corrections, or preferences. Classifier- and dynamics-guided methods~\citep{dynaguide,lpb} use latent visual dynamics models~\citep{dynamics_model,daydreamer} to steer generated actions toward desired outcomes. Our method extends contrastive energy verification beyond post-hoc candidate ranking to enable inference-time steering. Figure~\ref{fig:compare} summarizes the relationship between our method and related works.

\section{Preliminaries}
\textbf{Flow-matching-based VLAs.}
Flow-matching policies~\citep{flow_matching} generate actions by learning a continuous-time velocity field that transports samples from a simple prior to an action distribution. In flow-matching VLAs, the VLM backbone first encodes the current observation context into a conditioning representation $\mathcal{K}$. Using $\mathcal{K}$, the action expert transforms an initial noise sample $\mathbf{x}_0 \sim \mathcal{N}(\mathbf{0}, \mathbf{I})$ into an action chunk by integrating the learned velocity field $\mathbf{v}_\pi$ over normalized flow time $s$. With $N$ Euler steps and $s \in \{0,\frac{1}{N},\ldots,\frac{N-1}{N}\}$, the update is
\begin{equation}
    \mathbf{x}_{s + \frac{1}{N}}=\mathbf{x}_s + \frac{1}{N} \mathbf{v}_\pi(\mathbf{x}_s, s \mid \mathcal{K}),
    \label{eq:euler}
\end{equation}
where integration starts from $\mathbf{x}_0$ and terminates at $\mathbf{x}_1$. 

\noindent\textbf{Contrastive learning and energy-based models.}
Contrastive learning trains a scoring function to assign higher compatibility to positive samples than to negative ones. Energy-based models (EBMs)~\citep{energy_model} provide a natural formulation for this comparison by assigning each input $\mathbf{z}$ a scalar energy $E_{\theta}(\mathbf{z}) \in \mathbb{R}$, where lower energy indicates higher compatibility. Given context $\mathbf{c}$, the energy is written as $E_{\theta}(\mathbf{z}\mid\mathbf{c})$.
One common objective for learning such relative preferences is the margin-based ranking loss~\citep{contrastive_loss}:
\begin{equation}
    \mathcal{L}_c
    =
    \bigl[
        m
        +
        E_{\theta}(\mathbf{z}^{+}\mid\mathbf{c})
        -
        E_{\theta}(\mathbf{z}^{-}\mid\mathbf{c})
    \bigr]_+,
    \label{eq:margin_prelim}
\end{equation}
where $\mathbf{z}^{+}$ and $\mathbf{z}^{-}$ denote positive and negative samples under the same context $\mathbf{c}$, $m>0$ is the margin, and $[\cdot]_+=\max(0,\cdot)$. The loss encourages positives to have lower energy than negatives by at least $m$. We therefore formulate action verification as contrastive energy modeling, leveraging the fact that candidate selection requires only relative ranking rather than calibrated absolute rewards.

\begin{figure}[t]
    \centering
    \includegraphics[width=1.0\linewidth, trim=265 115 370 65, clip]{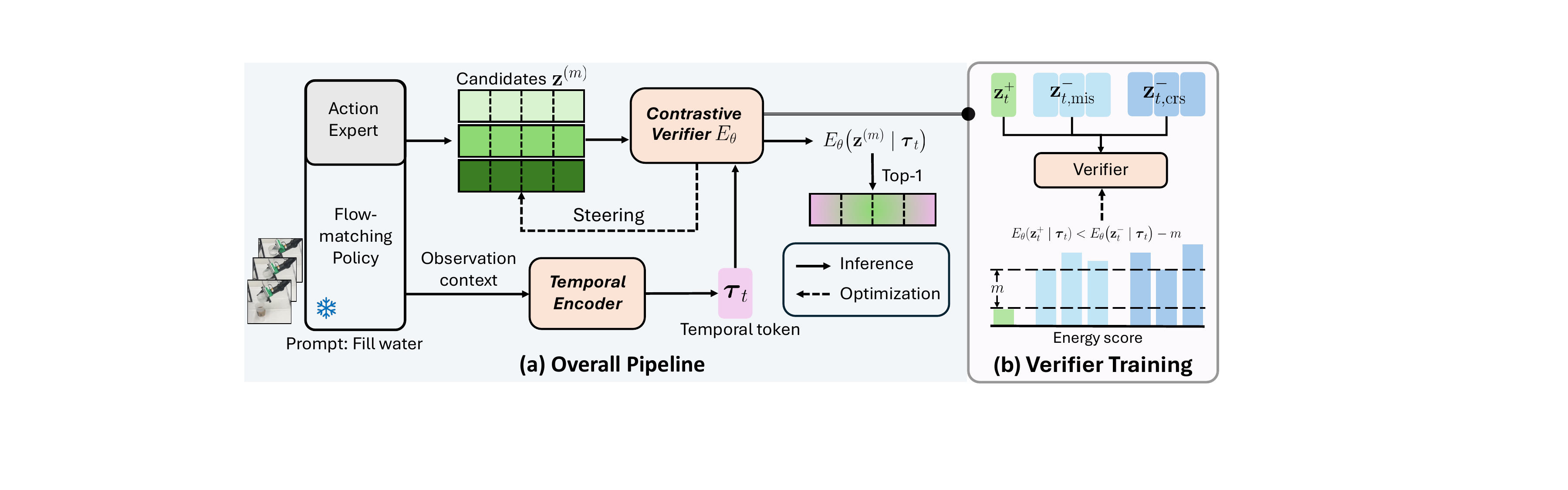}
    \vspace{-12pt}
    \captionsetup{
    justification=raggedright,
    singlelinecheck=false
    }
    \caption{\textbf{Illustration of Temporal Verification.}
    \textbf{(a)} Overall pipeline. We encode observation-action history into a temporal token and use it to condition the energy-based contrastive verifier. At each flow-matching integration step, candidate representations are scored by the verifier. The resulting energy scores are used to guide intermediate samples during integration and rank completed chunks after integration.
    \textbf{(b)} Verifier training process. Expert chunks serve as positives, while temporally mismatched chunks and policy-generated deviations serve as negatives. The verifier is trained to assign positives lower energy than negatives by at least a specified margin.
    }
    \label{fig:pipeline}
\end{figure}

\section{Methodology}
\textbf{Overview.}
Our goal is to augment a pretrained flow-matching VLA with a lightweight verifier that accounts for cross-chunk temporal coherence without requiring additional expert demonstrations.
To this end, we propose Temporal Verification (TeV), illustrated in Figure~\ref{fig:pipeline}(a). TeV first encodes recent observation--action history into a compact \emph{temporal token}, trained with a dynamics-aware objective to preserve execution-relevant temporal information. Using this token as temporal context, we train a lightweight energy-based verifier contrastively (Figure~\ref{fig:pipeline}(b)). At inference time, the learned energy function provides gradient-based guidance that steers intermediate flow samples toward lower-energy regions during generation, and then ranks the resulting action candidates for execution.

\subsection{Temporal Token Learning}
Augmenting stateless flow-matching VLAs with dense observation histories can incur substantial computational overhead~\citep{mem}. We instead compress recent history into a compact \textit{temporal token}, providing the verifier with temporal context at low inference overhead.
Given a history window of length $W$, we first construct a compact representation for each timestep $i \in \{t-W+1,\ldots,t\}$. 
For each observation $\mathbf{o}_i$, we extract its token representations from the pretrained VLM backbone. We project these tokens into a $d_h$-dimensional latent space and append a learnable readout token shared across timesteps. A lightweight transformer processes the resulting sequence with bidirectional attention, and its readout output forms the compressed observation summary $\mathbf{h}_i \in \mathbb{R}^{d_h}$. These representations compactly encode individual observations but remain temporally independent, and therefore do not yet capture how the trajectory evolves over time.

To incorporate execution history, we pair each observation summary $\mathbf{h}_i$ with the immediately preceding executed action $\mathbf{a}_{i-1}$. We concatenate each pair and process the resulting sequence with a causal temporal transformer $\mathcal{T}_\psi$~\citep{attention,decision_transformer}:
\begin{equation}
    \left(\boldsymbol{\tau}_{t-W+1}, \dots, \boldsymbol{\tau}_t\right)
    =
    \mathcal{T}_{\psi}
    \!\left(
        \left[\mathbf{h}_{t-W+1} ; \mathbf{a}_{t-W}\right], \dots,
        \left[\mathbf{h}_t ; \mathbf{a}_{t-1}\right]
    \right).
    \label{eq:temporal_token}
\end{equation}
The final output $\boldsymbol{\tau}_t$, termed the \textbf{temporal token}, causally summarizes recent observation--action history and provides temporal context for verifying candidate chunks at time $t$.

To encourage the temporal token to capture coherent observation--action evolution, we train it with an auxiliary future-action prediction objective. Let $g(\cdot)$ denote a lightweight prediction head that maps $\boldsymbol{\tau}_t$ to the next $F$ actions. We define
\begin{equation}
\mathcal{L}_{\tau} =
\left\|
g(\boldsymbol{\tau}_t) -
[\mathbf{a}_{t}, \ldots, \mathbf{a}_{t+F-1}]
\right\|^2,
\label{eq:temporal_loss}
\end{equation}
where $[\mathbf{a}_{t}, \ldots, \mathbf{a}_{t+F-1}]$ denotes the ground-truth future action sequence from the offline trajectory. This auxiliary objective serves as a dynamics-aware regularizer, encouraging $\boldsymbol{\tau}_t$ to retain temporally predictive information that connects recent observations and executed actions to subsequent motion. The resulting representation is thus better suited to temporally aware candidate ranking and steering.

\subsection{Contrastive Verifier Training}
We next introduce a lightweight energy model $E_{\theta}$ as the temporally aware verifier. Conditioned on the temporal token $\boldsymbol{\tau}_t$, it assigns an energy score to each sampled candidate action chunk under the recent execution context. Because candidate selection requires only relative ranking rather than calibrated reward estimation, we formulate verifier learning as a contrastive problem that assigns lower energy to expert chunks than to undesirable alternatives.

For each training timestep $t$, let $\mathbf{z}_t^{+}$ denote the latent representation of the corresponding expert chunk, which serves as the positive sample using supervision already available in the offline imitation data. We then construct two classes of negatives, each targeting a distinct failure mode of generative action prediction. Contrasting the expert chunk against these negatives trains the verifier to assess the action quality and temporal compatibility required for candidate ranking and generation-time guidance.

\noindent\raisebox{-0.6pt}{\ding[1.1]{182\relax}} \textbf{Temporal-mismatch negatives.}
Flow-matching policies can model multimodal action distributions, but this flexibility may cause switching between individually plausible yet mutually inconsistent action modes across consecutive chunks, leading to oscillatory behavior and task failure.
We therefore construct \emph{temporal-mismatch negatives} 
$\{\mathbf{z}_{t, \mathrm{mis},k}^{-}\}_{k=1}^{K_{\mathrm{mis}}}$ to train the verifier to recognize temporal compatibility. For each positive chunk $\mathbf{z}_t^{+}$, we sample expert chunks from other timesteps $j$ in the same episode subject to a minimum temporal distance $|j-t| \geq \delta$. Although these chunks originate from expert demonstrations and are therefore plausible in isolation, they are inconsistent with the execution history encoded by $\boldsymbol{\tau}_t$. Contrasting them with $\mathbf{z}_t^{+}$ encourages the verifier to distinguish the appropriate continuation of the current trajectory from plausible but temporally mismatched alternatives. This construction is particularly useful under state ambiguity, where similar observations may correspond to different stages of execution and therefore require different subsequent actions.

\noindent\raisebox{-0.6pt}{\ding[1.1]{183\relax}} \textbf{ Coarse-integration negatives.} 
Temporal-mismatch negatives capture expert-like chunks that are incompatible with the current history, but do not expose the verifier to quality degradation produced by the policy itself. We therefore construct \emph{coarse-integration negatives} 
$\{\mathbf{z}_{t, \mathrm{crs},k}^{-}\}_{k=1}^{K_{\mathrm{crs}}}$ from policy-generated chunks under the same observation context using fewer Euler steps $n<N$.
Reducing the number of integration steps yields a coarser approximation of the learned flow, providing a controlled source of imperfect policy-generated chunks. We sample negatives across different values of $n$, with smaller $n$ generally producing easier negatives and larger $n$ producing harder negatives that more closely resemble full-step policy outputs. Contrasting against these negatives encourages the verifier to distinguish high-quality action chunks from suboptimal policy-generated alternatives.

\begin{algorithm}[t]
\caption{Temporal Verification}
\label{alg:method}
\begin{algorithmic}[1]
\Require Velocity field $\mathbf{v}_{\pi}$, verifier $E_{\theta}$, temporal encoder $\mathcal{T}_{\psi}$
\Require Observation history $\{\mathbf{o}_{i}\}_{i=t-W+1}^{t}$, action history $\{\mathbf{a}_{i}\}_{i=t-W}^{t-1}$
\Require History length $W$, candidates $M$, Euler steps $N$, guidance schedule $\{\gamma_s\}$
\Ensure Selected action chunk $\mathbf{x}^{(m^*)}_1$
\vspace{2pt}
\State \textbf{// Encode temporal context}
\State Compute temporal token $\boldsymbol{\tau}_t$ via Eq.~\ref{eq:temporal_token}
\State Compute VLA context cache $\mathcal{K}_t$ from $\mathbf{o}_t$
\vspace{2pt}
\State \textbf{// Verifier-guided candidate generation}
\For{$m = 1, \ldots, M$}
    \State Sample $\mathbf{x}^{(m)}_0 \sim \mathcal{N}(\mathbf{0}, \mathbf{I})$
    \For{$j = 0, \ldots, N-1$}
        \State $s \gets j / N$
        \State Extract intermediate representation $\mathbf{z}^{(m)}_s$
        \State Update $\mathbf{x}^{(m)}_{s+1/N}$ via Eq.~\ref{eq:guided_update}
    \EndFor
    \State Compute final representation $\mathbf{z}^{(m)}$
\EndFor
\vspace{2pt}
\State \textbf{// Select best candidate}
\State $m^* \gets \argmin_{m} E_{\theta}(\mathbf{z}^{(m)} \mid \boldsymbol{\tau}_t)$ \Comment{Eq.~\ref{eq:verify}}
\State \Return $\mathbf{x}^{(m^*)}_1$
\end{algorithmic}
\end{algorithm}

Having defined the negative families, we train the verifier using the margin-based contrastive objective in Eq.~\ref{eq:margin_prelim}. Let $\mathcal{R} \coloneqq \{\mathrm{mis}, \mathrm{crs}\}$ denote the set of negative types. For each $r \in \mathcal{R}$, let $\{\mathbf{z}_{t,r,k}^{-}\}_{k=1}^{K_r}$ denote the corresponding negatives and and $m_r>0$ its margin. The contrastive verifier loss is
\begin{equation}
\mathcal{L}_c
=
\sum_{r \in \mathcal{R}}
\frac{1}{K_r}
\sum_{k=1}^{K_r}
\left[
m_r
+
E_{\theta}\!\left(\mathbf{z}_{t}^{+} \mid \boldsymbol{\tau}_t\right)
-
E_{\theta}\!\left(\mathbf{z}_{t,r,k}^{-} \mid \boldsymbol{\tau}_t\right)
\right]_+,
\label{eq:margin_refined}
\end{equation}
where $[x]_+ \coloneqq \max\{x,0\}$. Since lower energy indicates higher compatibility, minimizing $\mathcal{L}_c$ encourages the expert chunk $\mathbf{z}_t^+$ to receive lower energy than each negative sample by at least the corresponding margin $m_r$. Together, the two negative families provide complementary supervision for learning both temporal compatibility and action quality required for test-time verification.

Finally, we jointly optimize the temporal encoder and energy-based verifier with
\begin{equation}
    \mathcal{L}
    =
    \mathcal{L}_c
    \;+\;
    \lambda_{\tau} \mathcal{L}_{\tau},
    \label{eq:total}
\end{equation}
where $\lambda_{\tau}$ controls the contribution of the auxiliary future-action prediction objective.

\subsection{Verification as Guidance}
The learned verifier defines a differentiable energy function over action representations conditioned on recent execution history. This allows the learned energy to guide flow integration in addition to ranking completed candidates~\citep{guidance}. Let $\mathbf{z}_s$ denote the differentiable action representation extracted from the intermediate sample $\mathbf{x}_s$ at flow time $s$. Because $\mathbf{z}_s$ depends on $\mathbf{x}_s$, the energy gradient $\nabla_{\mathbf{x}_s} E_{\theta}(\mathbf{z}_s \mid \boldsymbol{\tau}_t)$ can be backpropagated through the action expert to guide generation. For each candidate $m$, we sample $\mathbf{x}^{(m)}_0 \sim \mathcal{N}(\mathbf{0}, \mathbf{I})$, while the VLA context $\mathcal{K}_t$ is computed once from the current observation and shared across candidates. We then augment the Euler update in Eq.~\ref{eq:euler} with an energy-gradient term:
\begin{equation}
    \mathbf{x}^{(m)}_{s+\frac{1}{N}}
    =
    \mathbf{x}^{(m)}_{s}
    +
    \frac{1}{N} \,
    \mathbf{v}_{\pi}\!\left(\mathbf{x}^{(m)}_{s}, s \mid \mathcal{K}_t\right)
    -
    \frac{\eta_s}{N}
    \nabla_{\mathbf{x}^{(m)}_{s}}
    E_{\theta}\!\left(\mathbf{z}^{(m)}_{s} \mid \boldsymbol{\tau}_t\right),
\label{eq:guided_update}
\end{equation}
where $\eta_s \geq 0$ controls the guidance strength at flow time $s$. We linearly decay the $\eta_s$ from $\eta_0=\eta$ to zero over the integration process, applying stronger guidance early in the flow and progressively recovering the unguided base dynamics near completion. This schedule allows the verifier to influence coarse action structure early while preserving the base policy's fine-grained refinement in later steps.

After integration, each candidate produces a final action chunk $\mathbf{x}^{(m)}_1$ with representation $\mathbf{z}^{(m)}_1$. We select the candidate with minimum verifier energy,
\begin{equation}
    m^{*} = \argmin_{m \in \{1,\dots,M\}} 
    E_{\theta}\!\left(\mathbf{z}^{(m)}_1 \mid \boldsymbol{\tau}_t\right),
\label{eq:verify}
\end{equation}
and execute $\mathbf{x}^{(m^*)}_1$. The verifier thus serves a dual role, guiding intermediate samples toward lower-energy regions during integration and ranking completed candidates for execution. Algorithm~\ref{alg:method} summarizes the full procedure.

\begin{table}[t]
\small
\centering
\caption{\textbf{Performance (\%) comparison on LIBERO-Plus.} Best performance in \textbf{bold}.}
\vspace{-4pt}
\adjustbox{width=1.0\linewidth}{
\begin{tabular}{l|ccccccc|c}
\toprule
\textbf{Method} & \texttt{Camera} & \texttt{Robot} & \texttt{Language} & \texttt{Light} & \texttt{Background} & \texttt{Noise} & \texttt{Layout} & Average \\
\midrule
Base & 41.8 & 72.3 & 79.9 & 82.8 & 84.4 & 76.2 & 84.9 & 73.2 \\
Random & 43.9 & 70.2 & 86.7 & 81.8 & 85.5 & 79.3 & 81.1 & 74.3 \\
TE & 44.4 & 72.3 & 82.2 & 76.3 & 85.1 & 74.8 & 84.0 & 73.0 \\
BID & 44.2 & 73.3 & 76.2 & 81.4 & 80.3 & 76.6 & 81.7 & 72.2 \\
V-GPS & 39.9 & 67.4 & 77.8 & 74.5 & 77.2 & 68.8 & 73.1 & 67.2 \\
LPB & 33.9 & 56.5 & 63.7 & 64.9 & 58.5 & 62.5 & 68.0 & 58.5 \\
DynaGuide & 33.9 & 59.5 & 61.9 & 65.0 & 61.2 & 58.8 & 70.2 & 58.6 \\
TACO & 44.9 & \textbf{77.1} & 90.1 & 77.4 & 86.2 & 77.1 & 87.2 & 76.0 \\
TACO$^{*}$ & 42.7 & 74.8 & 83.8 & 76.6 & 76.8 & 72.2 & 80.1 & 71.5 \\
FOREWARN & 43.4 & 73.3 & 87.5 & 80.7 & 83.0 & 78.6 & 86.2 & 75.0 \\
\midrule
\textbf{Ours} & \textbf{48.4} & 75.6 & \textbf{91.9} & \textbf{84.7} & \textbf{95.8} & \textbf{80.0} & \textbf{92.6} & \textbf{79.8} \\
\bottomrule
\end{tabular}
}
\label{tab:plus}
\end{table}

\section{Experiments}
We evaluate TeV in both simulation and real-world settings across diverse perturbations and manipulation tasks. In simulation, we follow the standard evaluation protocol for each benchmark. In addition to comparisons with baseline methods, we conduct diagnostic analyses to isolate the contribution of each component and better understand the mechanisms behind the observed performance gains.

\subsection{Simulation Environment} 
We evaluate on two simulation benchmarks: LIBERO-Plus~\citep{liberoplus} and RoboTwin 2.0~\citep{robotwin2}. LIBERO-Plus spans seven perturbation types: object layout (\texttt{Layout}), camera viewpoint (\texttt{Camera}), robot initial state (\texttt{Robot}), language instruction (\texttt{Language}), lighting condition (\texttt{Light}), background texture (\texttt{Background}), and sensor noise (\texttt{Noise}). We focus on the most challenging LIBERO10 suite, comprising 2,519 task instances.
For RoboTwin 2.0, we select seven representative manipulation tasks: 
\textit{adjust bottle position} (\texttt{Adjust Bottle}), 
\textit{pick dual bottles} (\texttt{Pick Bottles}), 
\textit{place container onto plate} (\texttt{Place Container}), 
\textit{stack two bowls} (\texttt{Stack Bowls}), 
\textit{place empty cup on coaster} (\texttt{Place Cup}), 
\textit{open laptop} (\texttt{Open Laptop}), and 
\textit{press stapler} (\texttt{Press Stapler}). 
For each task, we collect 50 expert demonstrations for training and evaluate the policy over 100 independent trials.

\noindent\textbf{Baselines.}
We compare TeV with verification-based and temporal-consistency baselines using the same $\pi_{0.5}$~\citep{pi05} backbone and checkpoint. Unless otherwise noted, all test-time scaling methods sample $M=4$ action chunks per observation.
\begin{itemize}[leftmargin=10pt,topsep=-4pt,itemsep=1pt,partopsep=1pt,parsep=1pt]
    \item \textbf{Random selection}: This baseline randomly selects one candidate chunk from the sampled set for execution, isolating the effect of multi-candidate sampling without any verification signal.
    \item \textbf{Temporal Ensembling (TE)}~\citep{aloha}: This baseline maintains a buffer of previously predicted action chunks and executes a weighted average of the actions assigned to the current timestep, smoothing execution across overlapping chunks and enhances temporal consistency.
    \item \textbf{Bidirectional Decoding (BID)}~\citep{bid}: BID samples multiple candidate chunks and selects the one most similar with the previously executed chunk over overlapping timesteps.
    \item \textbf{TACO}~\citep{taco}: TACO uses a lightweight Coin-Flipping Network to estimate pseudo-counts over internal policy representations, and uses these estimates to prefer action chunks closer to high-density regions of the training distribution. We also report \textbf{TACO$^{*}$}, which uses $M=50$ candidates per observation, matching the default candidate budget in the original paper.
    \item \textbf{Other test-time improvement methods.} We additionally compare against broader test-time improvement approaches, including the value-based action reranking method V-GPS~\citep{steer_rl}, generative policy-steering methods LPB~\citep{lpb} and DynaGuide~\citep{dynaguide}, and the dynamics-model-based verifier FOREWARN~\citep{human_loop_2}.

\end{itemize}

\begin{table}[t]
\small
\centering
\caption{\textbf{Performance (\%) comparison on RoboTwin2.0 tasks.} Best performance in \textbf{bold}.}
\vspace{-4pt}
\begin{tabular}{l|ccccccc|c}
\toprule
\textbf{Method} & \makecell{\texttt{Adjust} \\ \texttt{Bottle}} & \makecell{\texttt{Pick} \\ \texttt{Bottles}} & \makecell{\texttt{Place} \\ \texttt{Container}} & \makecell{\texttt{Stack} \\ \texttt{Bowls}} & \makecell{\texttt{Place} \\ \texttt{Cup}} & \makecell{\texttt{Open} \\ \texttt{Laptop}} & \makecell{\texttt{Press} \\ \texttt{Stapler}} & Average \\
\midrule
Base & 86.0 & 48.0 & 86.0 & 89.0 & 78.0 & 75.0 & 44.0 & 72.3 \\
Random & 85.0 & 52.0 & 90.0 & 88.0 & 79.0 & 72.0 & 50.0 & 73.7 \\
TE & 87.0 & 43.0 & 86.0 & 88.0 & 82.0 & 74.0 & 47.0 & 72.4 \\
BID & 87.0 & 59.0 & 87.0 & 91.0 & 76.0 & 77.0 & 47.0 & 74.9 \\
TACO & 81.0 & 46.0 & 87.0 & 85.0 & 80.0 & 62.0 & 49.0 & 70.0 \\
TACO$^{*}$ & 85.0 & 32.0 & 85.0 & 87.0 & 80.0 & 67.0 & 52.0 & 69.7 \\
\midrule
\textbf{Ours} & \textbf{95.0} & \textbf{64.0} & \textbf{94.0} & \textbf{93.0} & \textbf{88.0} & \textbf{81.0} & \textbf{54.0} & \textbf{81.3} \\
\bottomrule
\end{tabular}
\label{tab:robotwin}
\end{table}

\noindent\textbf{Results.}
Tables~\ref{tab:plus} and~\ref{tab:robotwin} report the performance comparisons. Temporal Ensembling and BID show mixed effects, often performing similarly to the base policy and random selection. This suggests that directly applying existing temporal-consistency baselines to these simulation benchmarks is insufficient. TACO improves performance, whereas TACO$^*$ underperforms by a large margin, indicating that increasing the candidate budget does not necessarily lead to higher success rates.
By contrast, TeV delivers more consistent improvements across the evaluated task suites, achieving average gains of 3.8\% over the strongest baseline on LIBERO-Plus and 6.4\% on RoboTwin 2.0 subtasks. These results indicate that verifier design is critical for translating additional test-time samples into performance gains. 

\begin{wraptable}{r}{0.5\linewidth}
  \small                     
  \centering              
  \vspace{-12pt}                                      
  \caption{\textbf{Performance (\%) over different $M$.}}
  \vspace{-6pt}        
  \adjustbox{width=1.0\linewidth}{
  \begin{tabular}{l| *{5}{S}}
  \toprule                            
  \textbf{Task} & {$M{=}1$} & {$M{=}2$} & {$M{=}4$} & {$M{=}8$} & {$M{=}16$} \\ \midrule  
  LIBERO-Plus    & 55.9 & 62.0 & 79.8 & 79.8 & 77.5 \\ \midrule  
  Adjust Bottle  & 76.0 & 84.0 & 95.0 & 88.0 & 87.0 \\ \midrule  
  Press Stapler  & 52.0 & 48.0 & 54.0 & 57.0 & 58.0 \\ \bottomrule  \end{tabular}
  }
  \vspace{-8pt}            
\label{tab:candidate_number}                           
\end{wraptable}
The limited gains of temporal-consistency baselines may stem from the relatively static task setups and simplified physical constraints in simulation. 
Both TE and BID refine the current prediction using only consecutive action chunks. In these benchmarks, however, the predicted chunks are often already clean and high quality due to well-collected expert demonstrations. This leaves limited room for smoothing-based improvement and similarity-based chunk selection. TACO performs less effectively on RoboTwin 2.0, possibly due to the limited number of expert demonstrations used to train the base policy. With sparse training coverage, pseudo-count-based estimates of training-distribution support may become less reliable. In contrast, TeV incorporates recent observation-action history and trains the verifier contrastively, providing a richer context for candidate evaluation and broader supervision over plausible but mismatched action chunks.

\begin{figure}[t]
    \centering
    \begin{minipage}[c]{0.52\linewidth}
        \centering
        \includegraphics[width=\linewidth, trim=0 0 0 0, clip]{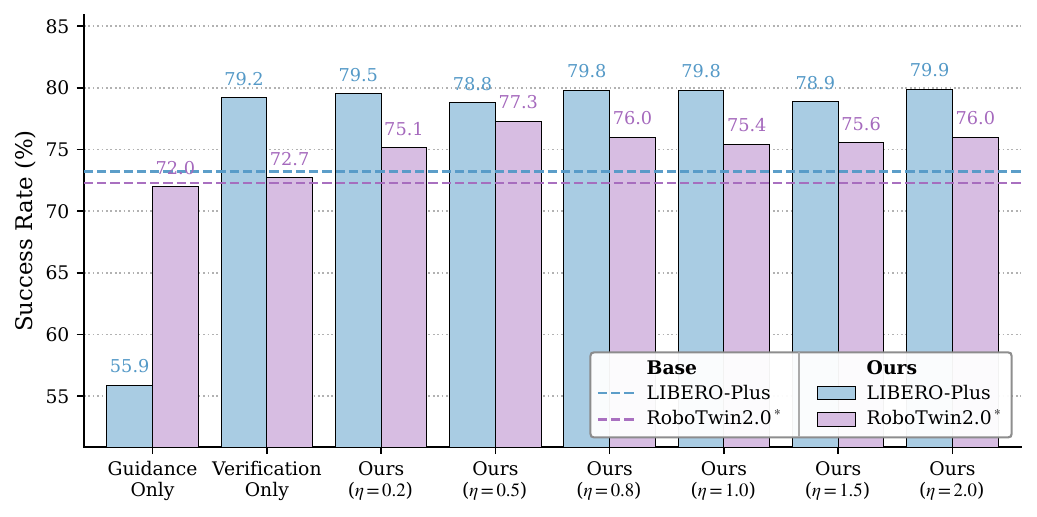}
        \vspace{-16pt}
        \captionof{figure}{\textbf{Effect of guidance and verification.}}
        \label{fig:abl_guide_verify}
    \end{minipage}
    \hfill
    \begin{minipage}[c]{0.465\linewidth}
        \centering
        \includegraphics[width=\linewidth, trim=0 0 0 10, clip]{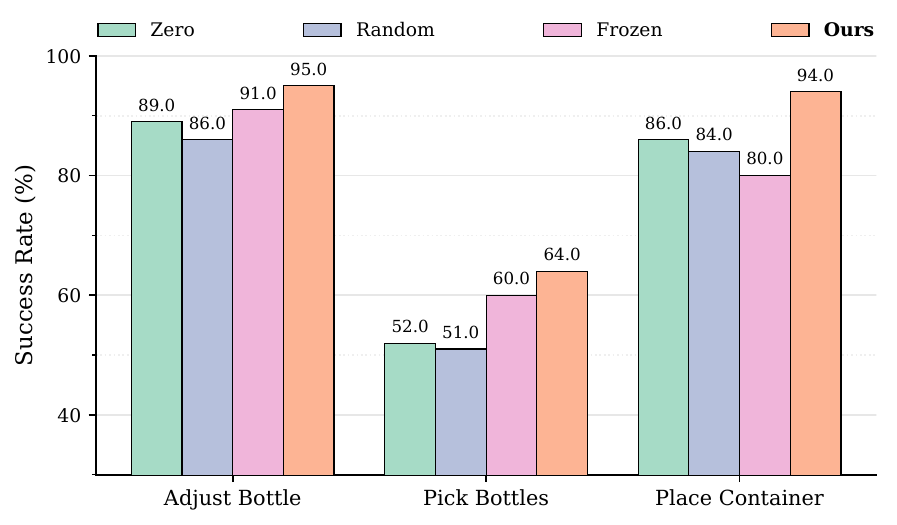}
        \vspace{-16pt}
        \captionof{figure}{\textbf{Effect of temporal token variants.}}
        \label{fig:abl_temporal}
    \end{minipage}
\end{figure}

\noindent\textbf{Ablations.}
We conduct the following ablation studies to isolate the contribution of individual components and assess the effect of key hyperparameters.
\begin{itemize}[leftmargin=10pt,topsep=2pt,itemsep=1pt,partopsep=0pt,parsep=0pt]
\item \textbf{Guidance and verification.} Figure~\ref{fig:abl_guide_verify} compares verification-only, guidance-only, and the full method across different guidance strengths \(\eta\). Verification-only consistently improves task success, whereas guidance alone can substantially degrade performance, indicating that gradient-based steering is unreliable without final candidate selection. The benefit of guidance is benchmark dependent. On RoboTwin2.0, adding guidance to verification improves success from 72.7\% to 75.1--77.3\% across \(\eta\in[0.2,2.0]\), while on LIBERO-Plus it provides little additional gain over verification alone. We attribute this difference to their distinct roles: verification selects among sampled candidates, whereas guidance refines intermediate samples during flow integration. Guidance is therefore most useful when the initial candidate set lacks a sufficiently accurate action, while final verification mitigates unfavorable steering. Overall, verification provides the robust core improvement, with guidance offering additional gains when candidate refinement is beneficial.

\item \textbf{Effect of negatives.} We ablate the two negative families on RoboTwin2.0. Starting from a 72.3\% base-policy success rate, using only coarse-integration negatives improves performance to 78.1\%, while temporal-mismatch negatives alone reach 77.4\%. Combining both further increases success to 81.3\%, indicating that the two negative families provide complementary training signals.

\item \textbf{Effect of the number of candidates.} Table~\ref{tab:candidate_number} examines how the candidate budget affects performance. Task success does not increase monotonically with the number of sampled candidates, indicating that additional samples are not automatically beneficial. This is expected, since an imperfect verifier may face harder ranking decisions as larger candidate sets include more out-of-distribution or low-quality chunks.

\item \textbf{Effect of the temporal token.} Figure~\ref{fig:abl_temporal} evaluates the learned temporal token against three variants: \textbf{\emph{Zero}}, which replaces it with $\mathbf{0}$; \textbf{\emph{Random}}, which replaces it with a Gaussian random vector; and \textbf{\emph{Frozen}}, which reuses the first temporal token computed in each episode for all subsequent steps. The learned temporal token consistently outperforms all three variants. Improvements over the \emph{Zero} and \emph{Random} baselines confirm the effectiveness of temporal-token conditioning, while the gain over \emph{Frozen} shows that dynamically updating the token with recent observation-action history provides meaningful context for verifier-based ranking and steering.
\end{itemize}

\subsection{Real-world Environment} 
\textbf{Setup.} 
We consider three tasks:
(1) put water bottle into the bag (\textit{Bottle to Bag}), 
(2) move the water cup from the shelf to the table (\textit{Cup Transport}), and 
(3) pour water from the right cup into the left cup (\textit{Water Filling}). 
We collect 180 trajectories in total for policy training. During execution, the robot runs at 15Hz with a fixed execution horizon. We compare TeV against the base policy, Bidirectional Decoding~\citep{bid}, and TACO~\citep{taco}. For execution safety, all methods adopt Temporal Ensembling~\citep{aloha}. Test-time scaling methods sample $M=4$ candidates per observation. We evaluate performance using a task completion score, assigning each episode an integer score according to the successfully completed substeps. Each setting is evaluated over 30 trials with varied object positions. 

\begin{figure}[t]
    \centering
    \begin{minipage}[c]{0.49\linewidth}
        \centering
        \small
        \captionof{table}{\textbf{Performance (\%) on real-world tasks.} Best results marked in \textbf{bold}.}
        \vspace{-6pt}
        \label{tab:real}
        \renewcommand{\arraystretch}{1.2}
        \resizebox{\linewidth}{!}{%
        \begin{tabular}{l|ccc}
            \toprule
            \textbf{Method} & \emph{Bottle to Bag} & \emph{Cup Transport} & \emph{Water Filling} \\
            \midrule
            Base & \val{60.7}{31.9} & \val{79.8}{27.5} & \val{40.5}{32.4} \\
            BID   & \val{59.8}{27.8} & \val{71.5}{25.9} & \val{51.3}{30.5} \\
            TACO & \val{64.0}{30.5} & \val{75.2}{21.2} & \val{51.5}{29.7} \\
            \midrule
            \textbf{Ours}     & \textbf{\val{69.3}{26.7}} & \textbf{\val{88.3}{15.9}} & \textbf{\val{68.0}{12.8}} \\
            \bottomrule
        \end{tabular}%
        }
    \end{minipage}
    \hfill
    \begin{minipage}[c]{0.5\linewidth}
        \centering
        \includegraphics[width=\linewidth, trim=0 0 0 0, clip]{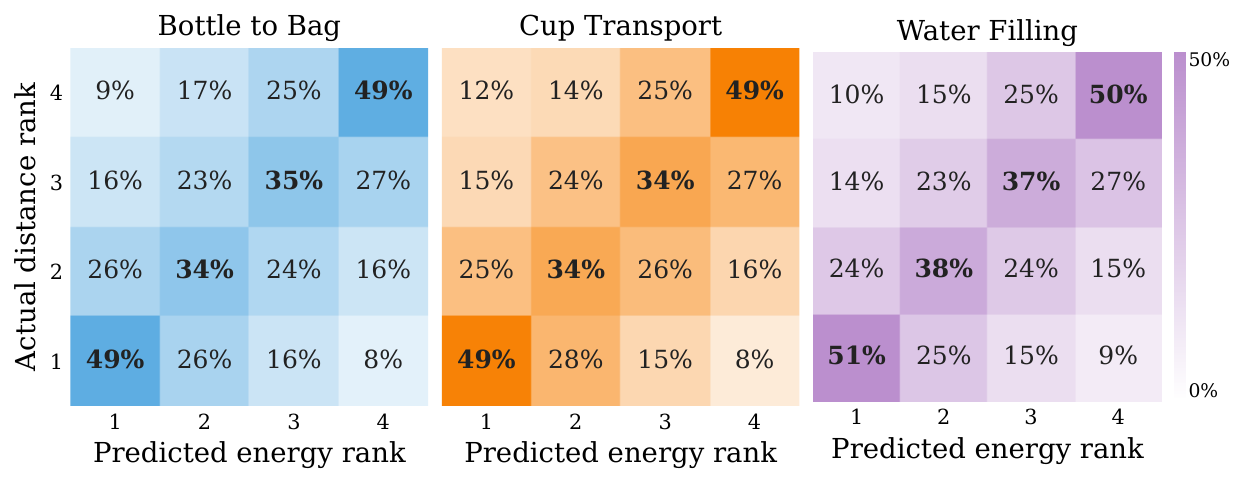}
        \vspace{-16pt}
        \captionof{figure}{\textbf{Energy score reliability.}}
        \label{fig:dis_heatmap}
    \end{minipage}
\end{figure}

\noindent\textbf{Results.}
Table~\ref{tab:real} reports the average task completion scores, including mean and standard deviation, across methods and tasks. Different from simulation, these real-world tasks require stable execution and smooth movement: the bottle in \emph{Bottle to Bag} is deformable, while the other two tasks involve transporting or pouring liquid. Across the baselines, we observe occasional trajectory oscillations and unstable execution, resulting in failures such as dropped bottles or spilled water. In contrast, TeV produces more stable and consistent trajectories, reducing failures associated with jerky motion and achieving stronger overall performance.
Baseline methods also sometimes terminate water pouring prematurely, before fully emptying the source cup. This may be due to partial observability during cup tilting, as neither the wrist camera nor the third-view camera reliably captures the liquid state inside the cups throughout the motion. By conditioning verification on encoded temporal history, TeV can use recent execution context as an additional signal of task progress, which may help maintain a consistent trajectory when the liquid state is only partially observable.

\noindent\textbf{Analysis.}
Figure~\ref{fig:dis_heatmap} evaluates the reliability of the predicted energy ranking. We compare the verifier-induced ranking of sampled candidates with their latent-space distance ranking to the corresponding expert action chunk. The verifier selects one of the two nearest candidates in more than 70\% of cases, indicating that lower-energy candidates generally align more closely with expert behavior. This supports the effectiveness of the contrastive verifier learning objective. 

\begin{figure}[t]
    \centering
    \includegraphics[width=1.0\linewidth, trim=150 15 275 20, clip]{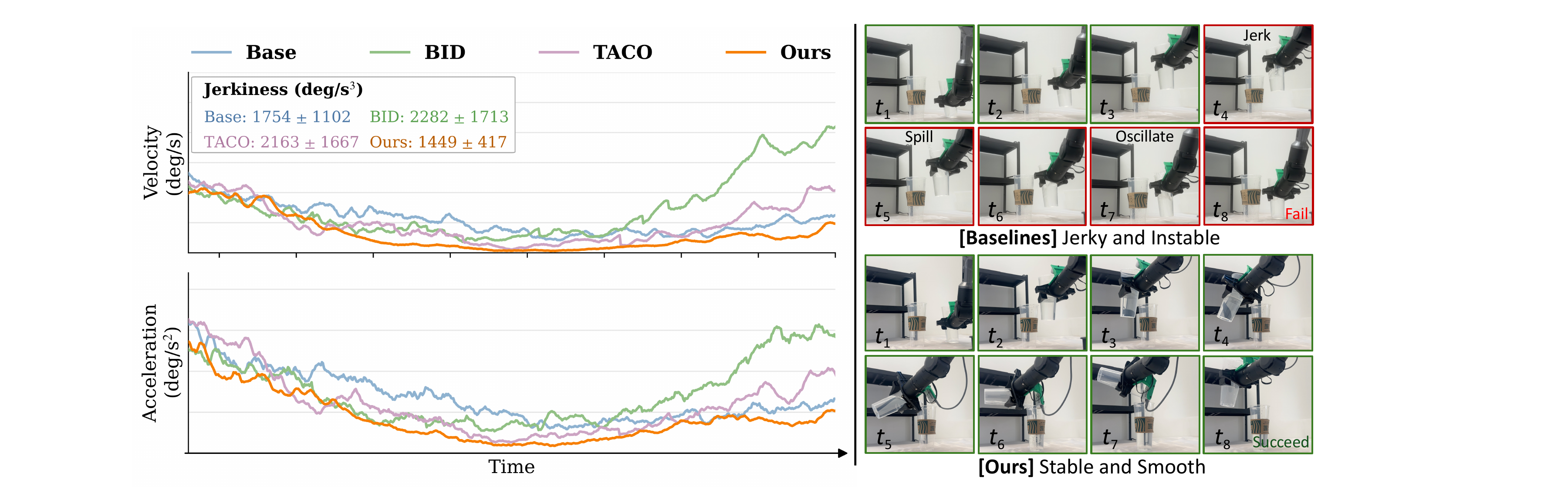}
    \vspace{-16pt}
    \captionsetup{
        justification=raggedright,
        singlelinecheck=false
    }
    \caption{\textbf{Illustration of robot kinematics and task progress over time.} \textbf{Left:} Average joint velocity and acceleration during interaction with the water cup. 
    \textbf{Right:} Real-world rollouts. The baselines often exhibit abrupt motions and oscillations, leading to water spilling or task failure, whereas our method maintains a steadier trajectory and completes the task without spilling.}
    \label{fig:kinematics}
\end{figure}

Figure~\ref{fig:kinematics} analyzes average robot kinematics and task progress rate. On the left, we illustrate metrics including joint velocity and acceleration, and quantify jerkiness as the per-trial mean magnitude of the third time derivative of joint positions, where lower values indicate fewer abrupt motion changes. The results show that TeV produces smoother kinematic profiles with lower jerkiness, which is associated with fewer failures such as water spilling. This supports its effectiveness in improving execution smoothness and temporal consistency. On the right, we provide example rollouts showing baseline failure cases, such as oscillatory motion and water spilling, which are not observed in the corresponding TeV rollouts. Additional rollout videos are provided in the supplementary.

\section{Conclusion}
Verification-based test-time scaling methods often suffer from temporal myopia, overlooking the consistency required across consecutive action chunks. In this paper, we introduced Temporal Verification (TeV), an efficient energy-based framework that makes action verification temporally aware.
TeV equips a lightweight energy-based verifier with a compact temporal token, enabling candidate actions to be evaluated against recent observation-action history rather than a static timestep. Since verification requires relative preference among candidates rather than calibrated reward estimation, we train the verifier with a contrastive objective.
The resulting verifier serves two roles: it ranks completed candidates for selection and actively steers intermediate policy outputs during generation. Extensive experiments in simulation and real-world manipulation tasks show that TeV is sample-efficient, improves success rates, and yields smoother robot dynamics.


\clearpage

\bibliographystyle{plainnat}
\bibliography{references}

\appendix
\renewcommand{\thefigure}{\thesection.\arabic{figure}}
\renewcommand{\thetable}{\thesection.\arabic{table}}

\section{Implementation Details}
\textbf{Training pipeline.}
We implement the temporal encoder as a lightweight transformer-style module with 3.0M parameters. The energy verifier is a shallow feedforward network with 0.9M parameters, equipped with spectral normalization that constrains the verifier's Lipschitz constant and stabilizes energy gradients. Together, the temporal encoder and verifier add less than 0.15\% parameters relative to the base model. For training-data construction, we uniformly sample from both negative families. The base model remains frozen throughout training, while only the added modules are optimized from scratch.

\noindent\textbf{Deployment framework.}
During deployment, we maintain a rolling buffer of recent observations and executed actions to compute the temporal token $\boldsymbol{\tau}_t$. At the first decision step, when no action history is available, we fall back to standard flow-matching generation; at later steps, any remaining missing action entries are zero-padded. To reduce test-time scaling latency, we use two optimizations. First, we compute the VLA key-value cache $\mathcal{K}_t$ once from the current observation and share it across all $M$ candidates, avoiding repeated backbone computation. Second, we use an efficient inference variant with \emph{guidance cutoff}. Verifier guidance is applied only during the first $\lfloor N/2 \rfloor$ Euler steps, after which candidates are scored and the $\lceil M/2 \rceil$ lowest-energy candidates are retained. The remaining integration steps are completed without guidance. This reduces gradient computation by half and decreases later-stage candidate updates with negligible accuracy degradation.

\section{Experimental Details}
\textbf{Inference acceleration.}
Test-time scaling with multiple sampled candidates increases inference latency, which can make real-world execution jerky and instable, leading to abrupt movements that degrade task performance. To reduce this overhead, we use two acceleration strategies. First, we reuse the VLA key-value cache across candidates, since all candidates are conditioned on the same visual-language context, and compile the action-sampling function to reduce execution overhead. Second, we apply progressive pruning: after the early guided integration steps, we retain only the lowest-energy candidates and complete the remaining integration with a smaller candidate set.s

Following~\citep{smolvla}, we measure latency as the elapsed time between observation acquisition and the availability of the corresponding action chunk in the control queue. On an RTX 5080 inference device with $M=4$ candidates, KV-cache reuse reduces latency by approximately 20--30 ms, compilation reduces latency by approximately 10 ms, and progressive pruning reduces latency by approximately 30--40 ms. For fair comparison, we apply the same applicable acceleration techniques to all baselines and to TeV, where progressive pruning is only used when the method supports candidate scoring during generation. Under the same RTX 5080, compilation and applicable cache optimizations, the measured latency ablation for TeV is shown in Figure~\ref{tab:latency}. Our real-world controller operates at 15 Hz, corresponding to 66.7 ms per action step. Using the discretized delay of $\lceil L/66.7 \rceil$, both the base and Verification-Only configurations incur a two-step control delay, whereas the guided configuration incurs a three-step delay. Thus, Verification-Only increases wall-clock latency but remains within the same discrete control-delay bucket as the base policy, while guidance adds one additional control step.

\begin{table}[h]
  \centering
  \caption{\textbf{Real-world latency analysis.}}
  \vspace{-4pt}
  \label{tab:latency}
  \adjustbox{max width=\linewidth}{%
  \begin{tabular}{lcccc}
    \toprule
    \textbf{Method} & \textbf{\textit{Bottle to Bag}} & \textbf{\textit{Cup Transport}} & \textbf{\textit{Water Filling}} & \textbf{Latency} $L$ \textbf{(ms)} \\
    \midrule
    Base                           & 60.7 $\pm$ 31.9 & 79.8 $\pm$ 27.5 & 40.5 $\pm$ 32.4 & \phantom{0}94.8 $\pm$ 0.1 \\
    Base + Verification            & 67.2 $\pm$ 28.1 & 87.2 $\pm$ 20.6 & 69.8 $\pm$ 14.2 & 118.9 $\pm$ 0.6 \\
    Base + Verification + Guidance & 69.3 $\pm$ 26.7 & 88.3 $\pm$ 15.9 & 68.0 $\pm$ 12.8 & 159.9 $\pm$ 1.5 \\
    \bottomrule
  \end{tabular}}
\end{table}

\noindent\textbf{Computational cost.} We train the verifier on RTX A6000 Ada GPUs. The training cost depends on the dataset size because negative construction is performed from the available offline trajectories. For the real-world tasks, verifier training takes approximately 0.6 GPU hours, which is significantly smaller compared to methods that rely on large-scaled datasets and models.

\end{document}